\documentclass[letterpaper, 10 pt, conference]{ieeeconf}  

\IEEEoverridecommandlockouts                              
\usepackage[utf8]{inputenc}
\usepackage[T1]{fontenc}
\usepackage{graphicx}
\usepackage{amssymb}
\usepackage{amsmath}

\title{\LARGE \bf
Facial Expression Editing with Continuous Emotion Labels
}
\author{\parbox{16cm}{\centering
 {\large Alexandra Lindt$^1$, Pablo Barros$^1$, Henrique Siqueira$^1$ and Stefan Wermter$^1$}\\
 {\normalsize
  $^1$ Knowledge Technology, University of Hamburg, Hamburg, Germany\\
 }}\newline \\ \\
 \parbox{16cm}{\centering
 {\large\textbf{ Pre-print version of \cite{us}} \\}}
 }

\begin{document}

\maketitle
\thispagestyle{empty}
\pagestyle{empty}

\begin{abstract}
Recently deep generative models have achieved impressive results in the field of automated facial expression editing. However, the approaches presented so far presume a discrete representation of human emotions and are therefore limited in the modelling of non-discrete emotional expressions. To overcome this limitation, we explore how continuous emotion representations can be used to control automated expression editing. We propose a deep generative model that can be used to manipulate facial expressions in facial images according to continuous two-dimensional emotion labels. One dimension represents an emotion's valence, the other represents its degree of arousal. We demonstrate the functionality of our model with a quantitative analysis using classifier networks as well as with a qualitative analysis.

\end{abstract}

\section{INTRODUCTION} 
Finding an automated way to alter the expression in a facial image is relevant for a variety of different fields, such as face animation, face recognition, emotion recognition or human-computer interaction. It has therefore been the topic of various publications over the last decade.
The most recent publications on the task
made use of deep generative models \cite{goodfellow_gan, vae, aae}. 
Radford et al. \cite{radford_2016} describe a variation of Generative Adversarial Networks (GAN) \cite{goodfellow_gan} named Deep Convolutional GAN (DCGAN), which can realize facial expression transfer through arithmetic operations on the latent input vectors of the generator network. However, the generated images are of low resolution and are limited to expressing only one discrete emotion class (happy, sad, excited, etc.).

Another recent solution, proposed by Yeh et al. \cite{yeh_autoenc_flow}, combines the structure of the Variational Autoencoder \cite{vae} with the approach of flow-based image warping \cite{yang_expression_flow}. The resulting Flow Variational Autoencoder (FVAE) is trained to create an expression flow map that can be used to transform an image to express a specific emotion. The FVAE is capable of generating high-resolution images and can even synthesize different intensities of a discrete emotion class in a face. Unfortunately, it can only be trained with paired data samples (i.e. pictures of the same face with different emotions), the emotion in the input image must be known and only images in which the face is shown frontally can be processed. 

The approach of Song et al., the Geometry-Guided GAN (G2-GAN) \cite{song_geometry_guided}, consists of two pairs of GAN \cite{goodfellow_gan} that form a mapping cycle in which one GAN applies an emotion to a neutral face image and the second GAN neutralizes the expressive face image. The emotion to be synthesized or removed is represented with facial geometry points, which makes the framework flexible in the emotions that can be synthesized and enables it to create different intensities of basic emotions. However, paired training data is necessary to train G2-GAN. 

Recently, Zhang et al. introduced the Conditional Adversarial Autoencoder (CAAE) \cite{zhang_aging} for manipulating the age of a face in an image. The model extends a conventional Adversarial Autoencoder \cite{aae} by a second discriminator network that ensures the generation of a photo-realistic output image. The CAAE is able to transform a facial image so that it corresponds to one out of ten discrete age groups. It does not require paired training data samples and has proven to be robust to variations in the input images \cite{zhang_aging}. 

Two approaches used the CAAE's structure as a basis.
The first one is the Conditional Difference Adversarial Autoencoder (CDAAE) \cite{cdaae}, which extends the CAAE by a feedforward connection between encoder and decoder network. Unfortunately, the CDAAE can only be trained with paired training data and the generated images have quite low resolution. The second CAAE-based approach is the Expression Generative Adversarial Network (ExprGAN) \cite{exprgan}. It extends the CAAE by an expression controller module that enables it to create discrete facial expressions of different intensities as well as by a face identity preserving loss function. 
Both ExprGan and CDAAE are able to synthesize mixed emotions as percentages of the emotion classes of their training set (e.g. 50\% sadness and 50\% fear).

Although very successful in their specific tasks, the approaches discussed above are in some ways limited in modelling non-discrete emotions because they presume a discrete representation of human emotions. This is probably due to the fact that until the recent release of the AffectNet database \cite{affectnet} only facial expression databases with discrete annotations were available. AffectNet's images, in contrast, are annotated with a two-dimensional vector that represents an depicted emotion as its degree of valence (unpleasant-pleasant) and arousal (relaxed-aroused). With the availability of this data, the motivation arises to investigate the applicability of continuous two-dimensional emotion representation in the field of automated expression editing.

With the goal of providing a controllable face expression editing mechanism that produces high-fidelity and high-quality facial translation, we employ the AffectNet database \cite{affectnet} to train a deep generative model for the manipulation of facial images according to continuous two-dimensional emotion labels.
To validate the contribution of our model to the generation of the desired emotional expressions, we conduct an objective experiment where individual neural networks are used to measure the arousal and valence of the generated images. Our experiments show that our model does edit the faces with the intended arousal and valence. Finally, we provide an analysis of how the proposed model imposes the facial expressions on the original images.

\section{Background}
\subsection{Representation of Human Emotion}
From a psychological point of view, there are two different approaches to categorizing human emotions and the corresponding facial expressions. A discrete or categorical representation of emotions assumes a set of fundamentally distinct basic emotions \cite{tomkins, ekman_basicemotions}. Since several scientists have defined these basic emotions differently \cite{tomkins, plutchik, izard_libero}, it is not clear which exact emotions belong to them. However, there is widespread agreement on the following six emotions: anger, disgust, fear, happiness, sadness and surprise \cite{peter_herborn}. In contrast, a continuous or dimensional emotion representation is based on the assumption that emotions cannot be divided into distinct groups, but can rather be described within a continuous space \cite{russell_circumplex, vastfjall}. Russell described emotions as points in the two-dimensional space of valence (unpleasant-pleasant) and arousal (relaxed-aroused) and termed this space \textit{The Circumplex Model of Affect} \cite{russell_circumplex}. 

There is no consensus on which model represents human emotions best. Several publications have investigated whether humans intuitively use a continuous or categorical representation of emotion. For both representation forms there are publications with evidence, some of them even contradict each other directly \cite{ekman_friesen_constants, carroll_russel}. Further, some researchers examined the connections between a perceived emotion and the physical state or language of a subject. More scientists were able to show clear correlations for the continuous model \cite{peter_herborn}. A detailed description and a list of all related experiments can be found in \cite{peter_herborn}.

\subsection{Deep Generative Models}
Generative modeling currently has two main approaches: Generative Adversarial Networks (GAN) \cite{goodfellow_gan} and Variational Autoencoders (VAE) \cite{vae}. GAN consist of a discriminator and a generator network that are trained simultaneously in a minimax two-player game. This process results in a generator network that is able to generate high-dimensional data samples similar to those of the training data set. An extension of GAN is the Conditional GAN (CGAN) \cite{cgan}, whose generated output is further influenced by a conditional variable. In contrast to GAN, the VAE is a stochastic model that consists of an encoder and a decoder network. The encoder network maps an input to a latent representation and the decoder network subsequently uses the latent vector to reconstruct the input. A prior distribution is imposed on the latent space through the loss function of the model. After training, the decoder network can be employed to create data samples from latent vectors that are sampled from the imposed distribution. The Adversarial Autoencoder (AAE) \cite{aae} integrates the idea of GAN into the VAE by using an adversarial training process to impose the prior distribution to the latent space. For this purpose, an additional discriminator network is employed as adversary. 

Our proposed model is a VAE with an additional discriminator network on the generated output. This additional network imposes the distribution of the training data on the generated data samples and, therefore, causes the generation of photo-realistic facial images that express the target emotion.

\section{Proposed Model}
For automated expression editing according to continuous two-dimensional emotion labels, we propose an updated version of the Conditional Adversarial Autoencoder (CAAE) \cite{zhang_aging}. We choose this model because it does not require paired data (i.e. pictures of the same face with different emotions) for training and has also proven to be robust against variations in the input images \cite{zhang_aging}. 

Inspired by ExprGan \cite{exprgan} and G2-GAN \cite{song_geometry_guided}, we extend the CAAE by an identity-preserving loss on the reconstructed image. This loss forces the output image to show the same person as the input image. The original CAAE approach trains encoder and generator network with a loss function on total variation minimization \cite{tv_loss} to reduce ghosting artifacts in the generated images. Since we have empirically found that these artifacts do not appear in the output images when the CAAE is trained with the identity-preserving loss, the total variation minimization loss is omitted in our approach. 
In order to improve the quality of the synthesized emotional expressions,  we further change the influence of the emotion label on the discriminator network on output and emotion label.

\begin{figure*}
 \includegraphics[width=0.9\textwidth]{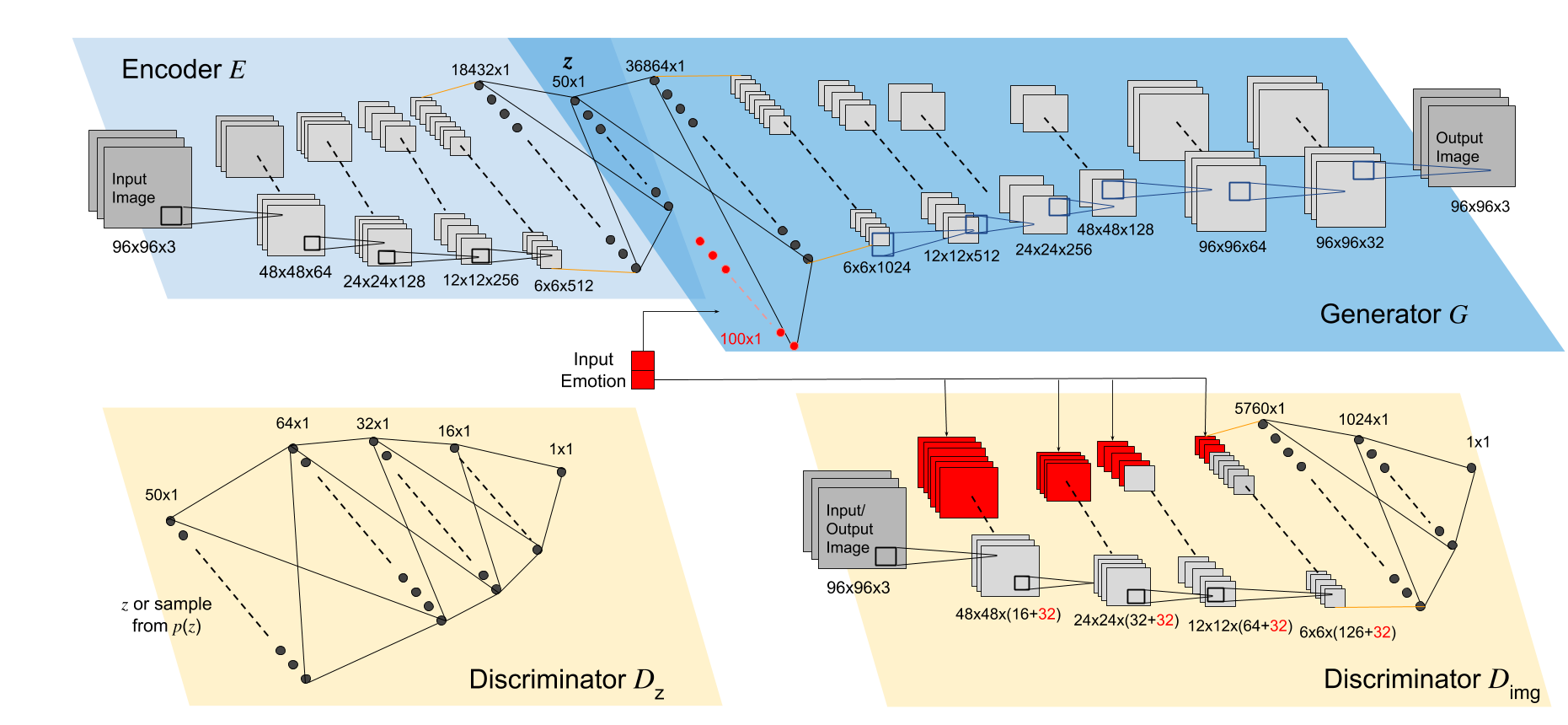}
 \caption{Detailed Architecture of the CAAE used for expression editing according to continuous two-dimensional emotion labels. The encoder network $E$ maps an input image $x$ to a smaller representation $z$ through four convolution layers and one fully connected layer. Note that the orange line denotes a reshape operation. The vector $z$ is concatenated to the enlarged input emotion label $y$ (bright red) and the resulting vector serves as input for the generator network $G$. $G$ consists of one fully connected layer followed by six transposed convolution layers and 
 outputs an image $x_{gen}$ of the same size as $x$ that shows the face from $x$ expressing $y$. The discriminator $D_z$ imposes the prior distribution $p(z)$ on the generated $z$ while the discriminator $D_{img}$ ensures that $x_{gen}$ is photo-realistic and expresses $y$.\label{abc}}
\end{figure*} 

\subsection{Architecture}
As illustrated in Fig. \ref{abc}, our model consists of a total of four networks: an encoder network \begin{math}E\end{math}, a generator network \begin{math}G\end{math} and two discriminator networks \begin{math}D_{img}\end{math} and \begin{math}D_{z}\end{math}. It receives a face image \begin{math}x\end{math} and a continuous two-dimensional emotional label \begin{math}y\end{math} as input and outputs an image $x_{gen}$ that displays the face from the input image expressing the emotion represented by \begin{math}y\end{math}. During the model's training, \begin{math}y\end{math} describes the emotion that is expressed in \begin{math}x\end{math}.

\subsubsection{Encoder and Generator}
Given an input image \begin{math}x\end{math} and respective emotion label \begin{math}y\end{math}, the encoder network \begin{math}E\end{math} first maps the input image to a lower-dimensional representation \begin{math}E(x)=z\end{math}, which represents the individual properties of the face depicted in input \begin{math}x\end{math}. This representation is subsequently used by the generator network \begin{math}G\end{math} to create a reconstruction \begin{math}G(z,y)=x_{gen}\end{math} of \begin{math}x\end{math} conditioned on \begin{math}z\end{math} and the two-dimensional input emotion label \begin{math}y\end{math}. 
In order to reconstruct \begin{math}x\end{math} as good as possible, both networks are trained with an image reconstruction loss \begin{math}L_{rec}\end{math}, which is defined in (\ref{caae_reconstruction_loss}). \begin{math}L_1\end{math} denotes the mean absolute error \cite{rmse}.
\begin{equation}
\label{caae_reconstruction_loss}
 \min_{E,G} L_{rec} = L_1 (x, G(E(x), z)) 
\end{equation}
We compute the identity-preserving loss in the same way as ExprGan \cite{exprgan} and therefore use the pre-trained VGG face model by Parkhi et al. \cite{Parkhi15}. The VGG face model originally classifies the identity of a face in an image. To compute whether it assumes two facial images to show the same person, the activations of five of its convolutional layers are compared. The calculation of the identity-preserving loss is defined as
\begin{equation}
\label{iden_loss}
 \min_{E,G} L_{iden} = \sum\limits_{l} L_1(\phi_{l}(x), \phi_{l}(G(E(x), z)).
\end{equation}
In this context, \begin{math}\phi_{l}\end{math} denotes the activation of the \begin{math}l\end{math}th layer. The considered layers are the \textit{conv1\_2}, \textit{conv2\_2}, \textit{conv3\_2}, \textit{conv4\_2} and \textit{conv5\_2} layer of the VGG face model.

\subsubsection{Discriminator $D_z$}
 The discriminator network \begin{math}D_z\end{math} is imposed on the encoder network's output \begin{math}z\end{math} and fosters it to be uniformly distributed. To this end, it receives either \begin{math}z = E(x)\end{math} or a sample from the uniform distribution \begin{math}p_{prior}(z)\end{math} as input and is trained to distinguish between both. The adversarial loss function between \begin{math}E\end{math} and \begin{math}D_z\end{math} is defined in (\ref{equ}). \begin{math}\mathbb{E}\end{math} denotes the likelihood \cite{Goodfellow-et-al-2016}, \begin{math}p_{prior}(z)\end{math} the prior distribution imposed on the internal representation \begin{math}z\end{math} and \begin{math}p_{data}(x)\end{math} the distribution of the training images.
\begin{equation} \label{equ}
\begin{aligned}
\min_E \max_{D_z} L_z = & \ \ \ \ \ \mathbb{E}_{z_{prior} \sim p_{prior}(z)}[\log D_z(z_{prior})] \\
 & + \ \mathbb{E}_{x \sim p_{data}(x)}[\log (1 - D_z(E(x)))] .
\end{aligned}
\end{equation}
\subsubsection{Discriminator $D_{img}$}
The second discriminator network \begin{math}D_{img}\end{math} is employed to ensure that the generator network \begin{math}G\end{math} produces a photo-realistic output image \begin{math}x_{gen}\end{math} that expresses the input emotion \begin{math}y\end{math}.
Therefore, \begin{math}D_{img}\end{math} receives the generated image \begin{math}x_{gen}=G(E(x),y)\end{math} and the input emotion label \begin{math}y\end{math} or an image and corresponding emotion label from the database as input. \begin{math}D_{img}\end{math} is trained to distinguish between generated and true pairs, while the generator network \begin{math}G\end{math} is trained to trick \begin{math}D_{img}\end{math} with its generated outputs. The adversarial loss function between \begin{math}G\end{math} and \begin{math}D_{img}\end{math} is defined in (\ref{g_d_img_loss}), where \begin{math}p_{data}(x,y)\end{math} denotes the distribution of the training data.

\begin{equation}\label{g_d_img_loss}
\begin{aligned}
\min_G \max_{D_{img}} L_{img} = & \ \ \ \ \mathbb{E}_{x,y \sim p_{data}(x,y)}[\log D_{img}(x,y)] \\
 & + \ \mathbb{E}_{x,y \sim p_{data}(x,y)} \\
& \ \ \ \ [\log (1 - D_{img}(G(E(x),y),y)] .
\end{aligned}
\end{equation}

\subsubsection{Overall Loss Function}
 The overall loss-function \begin{math}L_{total}\end{math} of the model is described in (\ref{total_loss}) as the weighted sum of all loss functions. The coefficients \begin{math}\lambda_1\end{math}, \begin{math}\lambda_2\end{math}, \begin{math}\lambda_3\end{math} and \begin{math}\lambda_4\end{math} balance the resolution of the generated images, the quality of the generated emotions and the obtained identity features in the generated images.
 \begin{equation} \label{total_loss}
 \min_{E, G} \max_{D_z, D_{img}} L_{total} = \lambda_1 L_{rec}+ \lambda_2 L_{iden} + \lambda_3 L_{z}+ \lambda_4 L_{img}
 \end{equation}

\subsection{Implementation Details}\label{hyperparams}
\subsubsection{Structural Details}

The following details extend on Fig. \ref{abc}. The encoder network \begin{math}E\end{math} is a Convolutional Neural Network \cite{lecun_nature} that consists of four convolution layers and one fully connected output layer. Inspired by DCGAN \cite{radford_2016}, a convolution of stride 2 is employed instead of pooling. This allows the encoder network to learn its own spatial downsampling \cite{zhang_aging}. 

\begin{math}E\end{math} obtains an image \begin{math}x\end{math} as input and transforms into a vector \begin{math}z\end{math} of size \begin{math}50\times1\end{math}. The two-dimensional input emotion label \begin{math}y\end{math} is scaled to 50 times its size and subsequently concatenated to \begin{math}z\end{math}. The resulting vector serves as input for the generator network \begin{math}G\end{math}, which up-samples its input via one fully connected layer and six transposed convolution layers \cite{transposed_conv} into an output image of the same size as the input image. The first four layers use a stride of 2, the last two layers use a stride of 1. All convolution layers in \begin{math}E\end{math} and transposed convolution layers in \begin{math}G\end{math} use a kernel size of 5. All values of \begin{math}x, y\end{math} and \begin{math}z\end{math} are in \begin{math}[-1,1]\end{math}.

\begin{math}D_{img}\end{math} consists of four convolution layers and two fully connected layers. The convolutions have a stride of 2 and kernel size of 5. Batch normalization \cite{lReLu2} is applied after each convolution. Following each block of convolution and batch normalization, the enlarged emotion label \begin{math}y\end{math} is concatenated to the block's output. We found this repeated concatenation of $y$ crucial for the performance of our CAAE, as it leads to a higher quality of the emotions synthesized in the output image \begin{math}x_{gen}\end{math}.
\begin{math}D_{img}\end{math} outputs a value from \begin{math}[0,1]\end{math}, which represents the estimated probability of its input being a pair of image and emotion label from the original data set.

Finally, \begin{math}D_{z}\end{math} receives a \begin{math}50\times1\end{math} vector as input that is either the output \begin{math}z\end{math} of the encoder network or a 50-dimensional vector sampled from the uniform distribution over \begin{math}[-1, 1]^{50}\end{math}. By using four fully connected layers, of which the first three are followed by batch normalization \cite{lReLu2}, the discriminator converts its 50-dimensional input into a one-dimensional output in \begin{math}[0, 1]\end{math} that represents the probability that \begin{math}D_{z}\end{math}'s input was sampled from the uniform distribution.

\subsubsection{Hyperparameters}
We trained the proposed model with the overall loss function defined in (\ref{total_loss}). The model has empirically been found to produce the best outputs for \begin{math}\lambda_1=1\end{math},
\begin{math}\lambda_2=\frac{1}{3}\end{math}, \begin{math}\lambda_3=0.01\end{math} and \begin{math}\lambda_3=0.01\end{math}. At the same time, the batch size should not be smaller than 49. To reduce computational cost, we only calculated the identity preserving loss for 16 out of 49 images in a batch.
The normal distribution with mean 0 and standard deviation 0.02 is employed for the initialization of the weights of all layers. All biases are initially set to 0.
For optimization, the Adam Optimizer \cite{adam} with learning rate \begin{math}\alpha=0.0002\end{math}, \begin{math}\beta_1=0.5\end{math} and \begin{math}\beta_2=0.999\end{math} is employed. We implemented the model using the machine learning framework TensorFlow \cite{tensorflow}.

Our particular architecture design and hyper parameters were found by using an empirical exploratory search. Our goal was to maximize the quality of the generated faces, and thus not represent an objective search. The observation of the generated faces were made purely subjective to the author's opinion, and yet, present an impressive performance on facial expression editing as demonstrated bellow.

\subsection{Facial Expression Editing}
 After training, our encoder and generator network can be employed to manipulate emotions in facial images according to arbitrary two-dimensional emotion labels. To modify an image \begin{math}x\end{math} to express an emotion \begin{math}y\end{math}, we feed \begin{math}x\end{math} to our trained encoder network \begin{math}E\end{math}, obtain the identity representation \begin{math}E(x)\end{math} and use the trained generator network \begin{math}G\end{math} to create an output image \begin{math}x_{gen} = G(E(x), y)\end{math}. The synthesized image \begin{math}x_{gen}\end{math} shows the face from \begin{math}x\end{math} expressing \begin{math}y\end{math}.

\section{Experimental Evaluation}
Following, we evaluate our model for its performance in automated expression editing according to continuous two-dimensional emotion labels. To enable an appropriate interpretation and analysis, we first explain three evaluation metrics. Then we take a closer look at the data set on which the model is trained. Finally, the capabilities of the model are investigated in two experiments.

\subsection{Evaluation Metrics}
\subsubsection{Root Mean Squared Error}\label{rmse}
The Root Mean Squared Error (RMSE) \cite{rmse} is a common evaluation metric for the difference between two vectors of the same size. Given two vectors \begin{math}x\end{math} and \begin{math}y\end{math}, both of size \begin{math}n\end{math}, it is defined as 
\[
RMSE(x,y) = \sqrt{\frac{1}{n} \sum_{i=1}^{n} (x_i-y_i)^2 }
\]

\subsubsection{Concordance Congruence Coefficent}\label{section_ccc}
The Concordance Congruence Coefficent (CCC) \cite{ccc} is a statistical measure of the agreement between the values of two equally sized vectors \begin{math}x\end{math} and \begin{math}y\end{math}. It combines the Pearson's correlation coefficient with the squared difference. 
The CCC is defined as 
\[
CCC(x,y) = \frac{2 \rho \sigma_x \sigma_y}{\sigma_x^2+ \sigma_y^2+(\mu_x - \mu_y)^2}
\]
where \begin{math}\rho\end{math} denotes the Pearson's correlation coefficient between the two vectors, \begin{math}\sigma^2\end{math} describes the variance of the respective vector and \begin{math}\mu\end{math} its mean value. The CCC can take values between -1 and 1 where 1 stands for a strong similarity and -1 for oppositeness. 
Unlike the Pearson Correlation Coefficient, the CCC penalizes predictions that are well correlated with the ground truth but shifted in value in proportion to their deviation. This property makes the CCC metric a meaningful metric for the evaluation of our two-dimensional emotion labels $\in [-1,1]^2$. Not only the correlation between the predicted emotion and the true emotion is considered, but also the prediction value's divergence from the real value.

\subsubsection{Sign Agreement}
The Sign Agreement (SAGR) \cite{sagr} is defined for two vectors \begin{math}x\end{math} and \begin{math}y\end{math} of equal length \begin{math}n\end{math} as
\[
SAGR(x,y) = \frac{1}{n} \sum_{i=1}^{n} \delta(sign(x_i), sign(y_i))
\]
where \begin{math}\delta(x,y) \end{math} denotes the Kronecker delta 
\[
 \delta(x,y) =
 \begin{cases}
 1 & \text{for } x = y \\
 0 & \text{for } x \neq x 
 \end{cases}
\]\noindent
The SAGR measures how much the signs of the individual values of two vectors \begin{math}x\end{math} and \begin{math}y\end{math} match. It takes on values in \begin{math}[0, 1]\end{math}, where 1 represents complete agreement and 0 represents complete contradiction. The SAGR metric is well suited to evaluate the predictions of our emotion labels \cite{affectnet}. Let's consider a face image that expresses a valence of +0.3 (i.e. a slightly positive valence). Although the two predictions +0.7 and -0.1 would have the same RMSE, the prediction of +0.7 is better suited, since +0.7 also denotes a positive valence \cite{affectnet}.

\subsection{Data Set}
The AffectNet database \cite{affectnet} is an extensive database of facial images that are annotated with both a categorical and a continuous two-dimensional emotion representation. The two-dimensional emotion labels are \begin{math}\in [-1, 1]^2\end{math}, where one dimension represents the emotion's valence (unpleasant-pleasant) and one represents its degree of arousal (relaxed-aroused). In the categorical representation of emotions, a distinction is made between eleven discrete emotion classes. The database contains about 450.000 manually annotated images. A further 550.000 included images were labeled with a classifier network trained on the hand-annotated images. The images were collected using search queries of emotion-related keywords in different languages in three major search engines. The pictures are therefore very diverse in terms of lighting, colors, camera angle and background as well as in head position, age, gender and ethnicity of the subjects. The facial expressions shown in the images are mostly natural and spontaneous. 

Although the images were labeled by skilled annotators, there are considerable variations in the annotations of different individuals. To measure the agreement between the valence and arousal values chosen by different annotators, 36.000 images were annotated by two people. Table \ref{tab:anno_agree} shows the annotations' agreement measured in RMSE, SAGR and CCC. It is apparent that there are noticeable differences in annotation, especially with regard to the arousal values.
\subsection{Experiments}
\subsubsection{Quantitative Experiment}
For a quantitative evaluation of the images generated by our model, we need a measure of the quality of the synthesized emotions. For this purpose, we train two Convolutional Neural Networks (CNN) \cite{lecun_nature} on the classification of valence and arousal in face images of size $96\times96\times3$, which is the size of images generated by our model. It should be noted that the analysis of valence and arousal in natural face images is not an easy task. As already demonstrated by table \ref{tab:anno_agree}, even human experts struggle to produce consistent annotations. 
Both classifier networks have the same architecture specified in table \ref{classifier_architecture}. 
To determine this structure, we trained a variety of networks on the hand-annotated AffectNet \cite{affectnet} images and selected the one with the best performance on our validation set of 4500 omitted AffectNet images. The final structure is oriented towards a CNN proposed for valence and
\bgroup
\def\arraystretch{1.5}
\begin{table}[htbp] 
\centering
\begin{tabular}{cc|c|c}
\cline{2-3}
& \multicolumn{1}{ |c| }{Valence} & \multicolumn{1}{ c| }{Arousal} \\ 
\cline{1-3}
\multicolumn{1}{ |c }{\textbf{RMSE}} &
\multicolumn{1}{ |c| } {0.340} & 0.362 \\ \cline{1-3} 
\cline{1-3}
\multicolumn{1}{ |c }{\textbf{SAGR}} &
\multicolumn{1}{ |c| } {0.815} & 0.667 \\ \cline{1-3} 
\cline{1-3}
\multicolumn{1}{ |c }{\textbf{CCC}} &
\multicolumn{1}{ |c| } {0.821} & 0.551 \\ \cline{1-3} 
\end{tabular}
\caption{Agreement between two annotators of the AffectNet database for valence and arousal labels respectively. Measured in RMSE, SAGR and CCC. \cite{affectnet} \label{tab:anno_agree}}
\end{table}
\egroup
arousal classification in \cite{cnn_dimensional_emotion_recognition}, which is itself based on the VGG-16 model \cite{vgg_16}. Both of our classifier networks output a value in \begin{math}[-1,1]\end{math} which expresses the estimated valence or arousal respectively. The classifiers were trained using stochastic gradient descent \cite{sgd} with learning rate 0.001 and a momentum of 0.9. The mean absolute error \cite{rmse} was employed as loss function. After training, our classifier networks achieve an accuracy that corresponds to that of other networks trained for the identical or similar task \cite{cnn_dimensional_emotion_recognition, affectnet, omg_challenge}. Table \ref{table:acc_class} presents the classifiers networks' accuracy on the validation data measured by the previously introduced evaluation metrics. For the quantitative evaluation of our model, we employ the classifier networks for the classification of 220500 images generated from our model. 
\subsubsection{Qualitative Experiment}
To investigate which facial attributes are changed by our model in order to make a face image express a particular emotion, we apply our model to 200 uniform face images for 49 different emotion labels each. 
The images used in this experiment
\def\arraystretch{1.5}
\begin{table}[htbp] 
\centering 
\begin{tabular}{|l|}
\hline
\textbf{Classifier Network Layer} \\\hline
 Convolution(Kernel 3x3, Stride 1, 64 Filters) \\ 
 Max-Pooling(Kernel 2x2, Stride 2) \\ 
\hline
Convolution(Kernel 3x3, Stride 1, 128 Filters) \\ 
Max-Pooling(Kernel 2x2, Stride 2) \\ 
\hline
Convolution(Kernel 3x3, Stride 1, 256 Filters) \\ 
Convolution(Kernel 3x3, Stride 1, 256 Filters) \\
Max-Pooling(Kernel 2x2, Stride 2) \\ 
\hline
Convolution(Kernel 3x3, Stride 1, 512 Filters) \\ 
Convolution(Kernel 3x3, Stride 1, 512 Filters) \\ 
Max-Pooling(Kernel 2x2, Stride 2) \\ 
\hline
 Fully Connected (4096 units) \\ 
 Dropout(dropout probability 0.5) \\ 
\hline
 Fully Connected (2622 units) \\ 
 Dropout(dropout probability 0.5) \\ 
\hline
 Fully Connected 2622 units) \\ 
 Dropout(dropout probability 0.5) \\ 
\hline
 Fully Connected (1 unit) \\ 
\hline
\end{tabular}
\caption{Detailed architecture of the classifier networks for the quantitative analysis}\label{classifier_architecture}
\end{table}
\begin{table}[htbp] 
\centering
\begin{tabular}{cc|c|c}
\cline{2-3}
& \multicolumn{1}{ |c| }{Classifier on Valence} & \multicolumn{1}{ c| }{Classifier on Arousal} \\ 
\cline{1-3}
\multicolumn{1}{ |c }{\textbf{RMSE}} &
\multicolumn{1}{ |c| } {0.450} & 0.411 \\ \cline{1-3} 
\cline{1-3}
\multicolumn{1}{ |c }{\textbf{SAGR}} &
\multicolumn{1}{ |c| } {0.676} & 0.708 \\ \cline{1-3} 
\cline{1-3}
\multicolumn{1}{ |c }{\textbf{CCC}} &
\multicolumn{1}{ |c| } {0.484} & 0.405 \\ \cline{1-3} 
\end{tabular}
\caption{Accuracy of the classifier networks employed for the quantitative evaluation of our model measured in RMSE, SAGR and CCC.\label{table:acc_class}}
\end{table}are taken from the CelebA data set \cite{celeb_a}, which contains portraits of prominent people. The images are aligned in such a way that the nose of the depicted person is located at the center of the image and the depicted faces are about the same size. We select the first 200 CelebA images that show a frontal face and cut them quadratically around their center. With this method, we obtain 200 quite precisely aligned facial images.

Each test image is processed by our trained network for 49 different two-dimensional emotion labels in \begin{math}[-1,1]^2\end{math}. One of the emotion labels is the neutral emotion [0,0] (i.e. neutral valence and arousal). Therefore, we receive an emotionally neutral version for every input image. Each of the 48 non-neutral generated images is compared with this neutral image. We deliberately choose to not compare the generated images with the input image, as the CelebA images are mainly red carpet images in which the subject laughs or smiles. 

For every non-neutral generated image, a grayscale heatmap that displays the differences to the corresponding neutral image is calculated. With \begin{math}R(p)\end{math} defining the red channel, \begin{math}G(p)\end{math} the green channel and \begin{math}B(p)\end{math} the blue channel of a pixel \begin{math}p\end{math} in in an image \begin{math}I\end{math}, the heatmap between two same-sized images \begin{math}I_1\end{math} and \begin{math}I_2\end{math} is computed pixel-wise for every \begin{math}p_1 \in I_1\end{math} and corresponding \begin{math}p_2 \in I_2\end{math} with 
\begin{align*}
max(&abs(R(p_1)-R(p_2)), \\
&abs(G(p_1)-G(p_2)), abs(B(p_1)-B(p_2))).
\end{align*}

\noindent
In the resulting heatmap, the brightest areas show the biggest differences between \begin{math}I_1\end{math} and \begin{math}I_2\end{math}. Since all faces in the test images have approximately the same position and the same size, we can add all \begin{math}200\end{math} heatmaps and normalize the outcome to get a general heatmap for each of the 48 non-neutral emotion labels.

\subsection{Results}
\subsubsection{Quantitative Analysis}
Table \ref{table:test_gen} shows the results of the evaluation of 220500 images generated by our model for various emotion labels. Of these 220500 images, 18000 express extreme emotions (i.e. very low/high arousal and very low/high valence). These extreme images are also evaluated separately, to enable a comparison between images generated for an average emotion label and those generated for extreme emotion labels.
The CCC values indicate a positive correlation between the emotion labels 
\begin{table}[htbp] 
\centering
\begin{tabular}{cc|c||c|c|c}
\cline{2-5}
& \multicolumn{2}{ |c|| }{All Images} & \multicolumn{2}{ c| }{Extreme Images} \\ 
\cline{2-5}
& \multicolumn{1}{ |c| }{Valence} & \multicolumn{1}{ c|| }{Arousal} & \multicolumn{1}{ c| }{Valence} & \multicolumn{1}{ c| }{Arousal}\\ 
\cline{1-5}
\multicolumn{1}{ |c }{\textbf{RMSE}} &
\multicolumn{1}{ |c| } {0.528} & 0.607 & 0.671 & 0.720\\ 
\cline{1-5}
\multicolumn{1}{ |c }{\textbf{SAGR}} &
\multicolumn{1}{ |c| } {0.567} & 0.483 & 0.691 & 0.624 \\
\cline{1-5}
\multicolumn{1}{ |c }{\textbf{CCC}} &
\multicolumn{1}{ |c| } {0.312} & 0.210 & 0.353 & 0.267 \\
\cline{1-5} 
\end{tabular}
\caption{Results of the classification of 220500 generated images (thereof 18000 extreme images) with our classifier networks. Masured in RMSE, SAGR and CCC.}\label{table:test_gen}
\end{table}
of the generated images and the emotion labels assigned to them by the classifier networks. It is higher for images generated for extreme emotions. Although the values do not indicate a strong similarity of the generated and classified labels,  the similarity can be assumed if one considers that this relatively low correspondence is also present among human annotators (see table \ref{tab:anno_agree}).
The SAGR metric shows that only half of the emotion label's signs match for all images. For extreme emotions, however, SAGR shows a clear connection between the signs of the emotion labels of the generated images and the emotion labels estimated by the classifiers. The RMSE is lower for images of all emotions than for images of extreme emotions. This suggests that for extreme images, larger errors were made in the emotion classification. When looking at the RMSE, however, it should be considered that the classifications for images with valence/arousal levels close to neutral cannot deviate as much from their true value as those for extreme emotions. To exemplify, the classification of the neutral emotion value 0 must be in [-1,1] and can, therefore, deviate by a maximum of 1 from the true value, while for an extreme emotion value (-1 or 1) the classified emotion can deviate up to 2. Furthermore, very large errors for single images (outliers) are heavily weighted in the RMSE \cite{rmse}.
\subsubsection{Qualitative Analysis}
\begin{figure}[thpb]
 \centering
 \includegraphics[scale=.36]{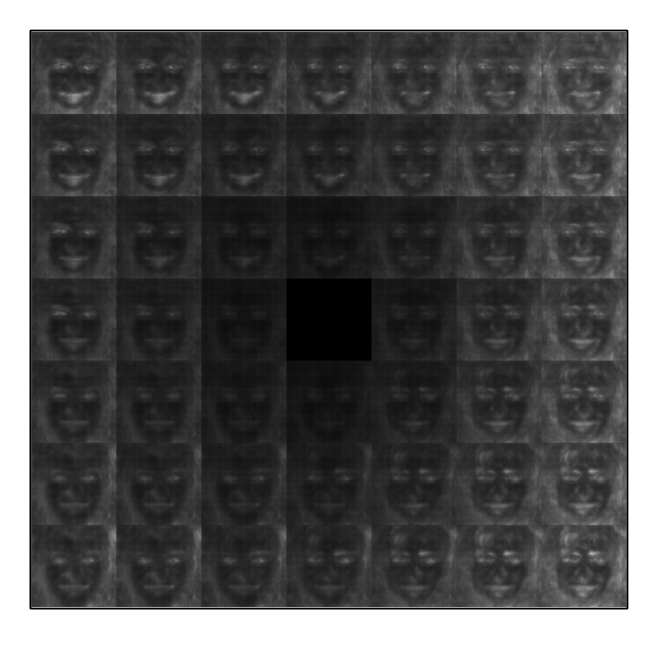}
 \caption{Computed overall heatmaps for 48 different emotion labels of the form [valence, arousal] $\in [-1,1]^2$, from high arousal (left) to low arousal (right) and from high valence (top) to low valence (bottom).
 The brightest areas in the heatmaps show the greatest overall differences between the images generated for the respective emotion label and the images generated for the neutral emotion label [0, 0]. See the digital version for details.}
 \label{result_qualitative}
\end{figure}

Fig. \ref{result_qualitative} shows the overall grayscale heatmaps for 48 different emotion labels.
Note that the changes in the images with weaker emotions (i.e. emotions close to the neutral emotion) are as expected smaller than in the images with at least one extreme emotion value (i.e. emotions of very high or low valence and/or very high or low arousal).
Furthermore, it can be noticed that there are strong changes in the background of the images. This background noise might have been caused by the fact that the only penalty the model receives for changing the background is the pixel wise reconstruction loss.
In contrast, changing the face in the right way is much more rewarding. 
Therefore, quite specific changes can be observed within the facial region. Our model primarily alters the mouth, eyebrows and eye area. 

For a high valence, the corners of the mouth are pulled upwards, the mouth is opened, the upper side of the eyes rises and a slight wrinkle forms under the eyes. These modifications might be associated with an emotion of positive value, such as happiness or delight. 
A lowered valence causes changes on the inside of the eyebrows as well as on the downside of the eye. These facial movements can be related to a depressed mood, where the inner sides of the eyebrows are moved down a little, pressing the eyes slightly downwards.

A high degree of arousal affects the upper lip, the upper part of the eyes and the rear eyebrow arch. All these changes are signs of an excited or surprised facial expression. Looking at the heatmaps for a low degree of arousal, there is clearly more background noise for which there is no discernible reason. However, within the facial region we can see changes in the area of the mouth as well as at the inner eyebrows and the eyes. These changes can be interpreted as signs of sleepiness, calmness or relaxation. The mouth becomes more closed, the eyes become narrower and the eyebrows move down a little.

Overall it can be found that the heatmaps combine the characteristics of their individual dimension values. For instance, the heatmap for a face with [high valence, high arousal] combines the heatmaps of [high valence, neutral arousal] and [neutral valence, high arousal]. Both high valence and high arousal cause a widening of the eyes. On the heatmap for combined high valence and high arousal, these changes seem to have added up to an even larger eye opening. Furthermore, the eyebrows are raised, just as it is characteristic for a high arousal value. The raised corners of the mouth, which were previously caused by high valence, can also be observed.

\section{Discussion, Conclusion And Future Work}

\subsection{Discussion}
\begin{figure}[thpb]
 \centering
 \includegraphics[scale=.29]{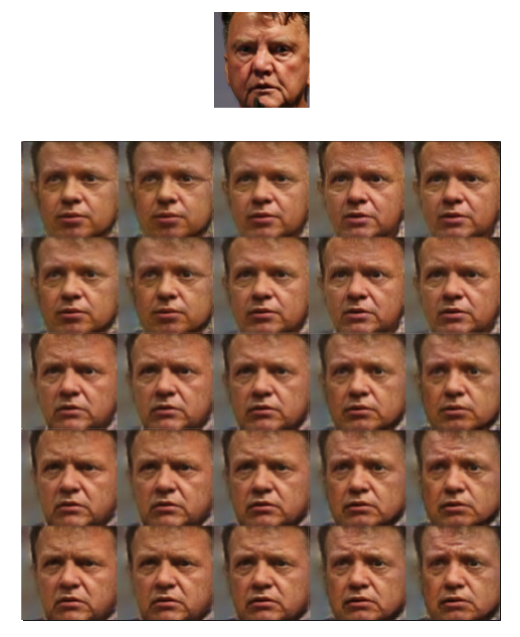}
 \caption{Examples of images generated by our model for one input image (top) and 25 different emotion labels $\in [-1,1]^2$, from high arousal (left) to low arousal (right) and from high valence (top) to low valence (bottom). See the digital version for details.
}
 \label{input_output}
\end{figure}

Our proposed model is well suited for editing facial expressions according to emotion labels from a two-dimensional emotion representation of valence and arousal. The generated images can be described as photo-realistic, the synthesized emotions appear natural and realistic and the identity of the person in the input image is largely maintained in the output image (see Fig. \ref{input_output}).
\begin{figure}[thpb]
 \centering
 \includegraphics[scale=.29]{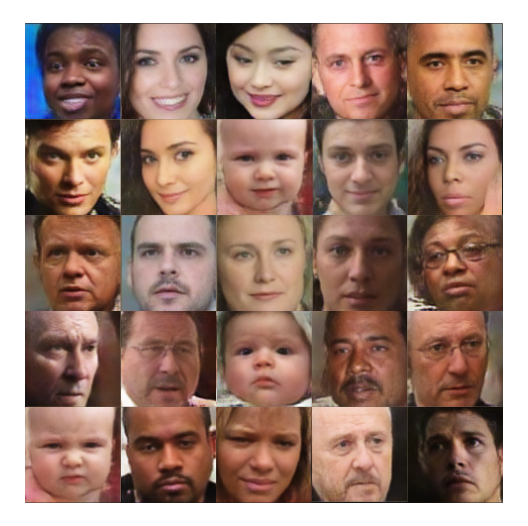}
 \caption{Examples of images generated by our model for 25 different emotion labels $\in [-1,1]^2$ and 25 different input images, from high arousal (left) to low arousal (right) and from high valence (top) to low valence (bottom). See the digital version for details.
}
 \label{examples}
\end{figure} 
 Our qualitative evaluation showed that the model ascribes certain characteristics to the values of the individual emotion dimensions and combines them sensibly for mixed emotion labels.
As depicted in Fig. 4, our model can be successfully applied to images that differ in terms of lighting, colors, the position of the depicted face or the characteristics of the subject, such as ethnicity, age or gender. The synthesized emotions are rather subtle, which is probably due to the fact that our training data contains almost only natural facial expressions which are in many cases less extreme than posed expressions or expressions observed in a laboratory environment for the reaction to emotional stimuli. 

The model produces overall consistent outputs. However,
our observations and the quantitative evaluation indicate
that extreme emotions in the input images can influence 
our model's output. The synthesized face images are less representative for their emotion label than the original face images and the emotions in images created for an extreme emotion label are better recognizable than in images created for an arbitrary emotion label. This suggest that a target emotion applied to an image changes its emotion only to a certain extent.
We further find that some images from our validation data set cannot be processed correctly by our model. These pictures typically have very low contrasts, an very unusual composition, coloring or illumination or parts of the depicted face are hidden or cut off.

\subsection{Conclusion}
In this paper, we introduced a neurocomputational model for facial expression editing according to continuous two-dimension emotion labels. The model is capable of generating high-quality and overall consistent outputs. From our observations we can conclude that our model changes faces in pictures in an understandable and plausible way. Our quantitative evaluation also demonstrates that our proposed model is able to generate facial expressions with continual conditions. The combination of the quantitative evaluation and the observations demonstrates how different conditions impact our solution, and how robust it is for different image conditions.

\subsection{Future Work}
Since our model can generate natural appearing facial images with realistic emotional expressions, its produced images could be used to train machine learning models for tasks like emotion recognition or face recognition. Our solution could be used as a boostrap for one-shot-based learning models. Also, the exploration of sequential data generation would be of fundamental importance for real-world applications.

\addtolength{\textheight}{-12cm}   




\section*{ACKNOWLEDGMENT}
The authors gratefully acknowledge partial support from the German Research Foundation DFG under project CML (TRR 169) and the European Union's Horizon 2020 research and innovation programme under the Marie Sklodowska-Curie grant agreement No 721619 (SOCRATES).


\end{document}